\documentclass[11pt,a4paper]{article}
\usepackage[utf8]{inputenc}
\usepackage{amsmath,amssymb}
\usepackage{graphicx}
\usepackage{hyperref}
\usepackage{cite}
\usepackage[margin=1in]{geometry}
\usepackage{booktabs}

\title{Emergent Lexical Semantics in Neural Language Models:\\ Testing Martin's Law on LLM-Generated Text}

\author{Kai Kugler \\ Trier University}

\date{}

\begin{document}

\maketitle

\begin{abstract}
We present the first systematic investigation of Martin's Law---the empirical relationship between word frequency and polysemy---in text generated by neural language models during training. Using DBSCAN clustering of contextualized embeddings as an operationalization of word senses, we analyze four Pythia models (70M--1B parameters) across 30 training checkpoints. Our results reveal a non-monotonic developmental trajectory: Martin's Law emerges around checkpoint 100, reaches peak correlation ($r > 0.6$) at checkpoint $10^4$, then degrades by checkpoint $10^5$. Smaller models (70M, 160M) experience catastrophic semantic collapse at late checkpoints, while larger models (410M, 1B) show graceful degradation. The frequency-specificity tradeoff remains stable ($r \approx -0.3$) across all models. These findings suggest that compliance with linguistic regularities in LLM-generated text is not monotonically increasing with training, but instead follows a balanced trajectory with an optimal semantic window. This work establishes a novel methodology for evaluating emergent linguistic structure in neural language models.
\end{abstract}

\section{Introduction}

Natural languages exhibit statistical regularities that reflect fundamental principles of communication and cognition. Among these, \textit{Martin's Law} describes the robust positive correlation between word frequency and polysemy (number of senses): high-frequency words tend to have more meanings \cite{martin1969specialization,zipf1949human}. This relationship has been documented across diverse languages and corpora, suggesting it emerges from pressures toward efficient communication \cite{piantadosi2012word}.

The advent of large language models (LLMs) trained on massive text corpora raises a fundamental question: Do these models generate text that exhibits the same linguistic regularities observed in human language? While recent work has examined scaling laws \cite{kaplan2020scaling}, syntactic emergence \cite{wei2022emergent}, and semantic capabilities \cite{bommasani2021opportunities}, the question of whether LLM-generated text respects \textit{lexical-semantic laws} like Martin's Law remains unexplored.

This gap is particularly important because Martin's Law reflects deep properties of how meaning is distributed in language. Testing whether this law emerges in LLM-generated text---and how it evolves during training---can reveal whether models develop human-like semantic organization or merely surface-level statistical mimicry.

\textbf{Our contributions:} We present the first investigation of Martin's Law in LLM-generated text, examining its emergence and evolution across training. Using contextualized embeddings and clustering-based sense identification, we analyze text generated by four Pythia models \cite{biderman2023pythia} at 30 checkpoints spanning the full training trajectory. We find that: (1) Martin's Law emerges early in training but peaks at an intermediate checkpoint before degrading; (2) smaller models experience catastrophic semantic collapse at late checkpoints; (3) the frequency-specificity tradeoff remains remarkably stable across training scales.

\section{Background}

\subsection{Martin's Law}

Martin's Law, first systematically described by Martin \cite{martin1969specialization} and presaged by Zipf \cite{zipf1949human}, states that word frequency and polysemy are positively correlated: $P(w) \propto f(w)^\beta$, where $P(w)$ is the number of senses (polysemy) of word $w$, $f(w)$ is its frequency, and $\beta \approx 0.5$--0.7 in natural language corpora.

This relationship has been explained through competing theoretical frameworks. The \textit{causal differentiation hypothesis} suggests that frequent words acquire more senses because they are used in more contexts \cite{zipf1949human}. The \textit{protectionist hypothesis} argues that polysemous words are easier to process and therefore increase in frequency \cite{adelman2006contextual}. Regardless of causal direction, the correlation itself is highly robust across languages \cite{koppen2007word}.

A related phenomenon is the \textit{frequency-specificity tradeoff}: high-frequency words tend to be semantically general (low specificity), while rare words are more specific \cite{piantadosi2012word}. This is typically measured as a negative correlation between frequency and semantic variance.

\subsection{Polysemy Measurement via Clustering}

Traditional approaches to measuring polysemy rely on dictionary sense inventories \cite{kilgarriff1997don}, which are subjective and unavailable for model-generated text. We adopt a data-driven approach: treating word senses as clusters in contextualized embedding space.

Given a set of contextualized embeddings $\{\mathbf{e}_i\}$ for all instances of word $w$, we apply DBSCAN clustering \cite{ester1996density} with cosine distance. The number of resulting clusters (excluding noise) serves as our polysemy estimate: $P(w) = |\text{clusters}(w)|$. This approach has been validated for sense induction tasks \cite{amrami2018word,schutze1998automatic}.

\section{Methods}

\subsection{Models and Data}

We analyze four models from the Pythia suite \cite{biderman2023pythia}: \texttt{pythia-70m}, \texttt{pythia-160m}, \texttt{pythia-410m}, and \texttt{pythia-1b}. These models are trained on the Pile dataset \cite{gao2020pile} with identical data order, enabling controlled comparison across scales.

For each model, we sample 30 checkpoints logarithmically spaced from initialization to final training step ($\sim 10^5$ steps). At each checkpoint, we generate 100 text samples (512 tokens each) using the model with temperature 1.0.

\subsection{Semantic Analysis Pipeline}

For each checkpoint:

\textbf{1. Embedding extraction:} We load the checkpoint and extract final-layer hidden states for all tokens in the generated samples. We filter to alphabetic tokens $\geq 3$ characters, excluding special tokens and punctuation.

\textbf{2. Polysemy computation:} For each word $w$ appearing $\geq 5$ times, we cluster its contextualized embeddings using DBSCAN ($\epsilon=0.3$, min\_samples=2, cosine metric). The number of clusters represents $P(w)$.

DBSCAN is particularly well-suited for sense induction because it does not require pre-specifying the number of clusters and can identify noise points (word uses that don't belong to any coherent sense cluster). However, the choice of clustering algorithm involves tradeoffs. Alternative approaches include:

\begin{itemize}
    \item \textbf{Agglomerative clustering:} Provides hierarchical sense structure but requires specifying the number of clusters per word, either via a global threshold or per-word heuristics.
    \item \textbf{Gaussian Mixture Models:} Can automatically determine cluster count via BIC/AIC model selection, but assumes spherical cluster shapes.
    \item \textbf{Affinity Propagation:} Automatically determines cluster count but is computationally expensive for large embedding sets.
\end{itemize}

We selected DBSCAN for its theoretical alignment with sense induction (not all words need to be polysemous) and computational efficiency. The $\epsilon$ parameter controls sense granularity: smaller values produce finer-grained senses, larger values merge related uses. Our choice of $\epsilon=0.3$ represents a moderate granularity, but sensitivity analysis across parameter values would strengthen future work.

\textbf{3. Specificity computation:} Semantic specificity is computed as the inverse of embedding variance: $S(w) = 1 / (\text{Var}(\mathbf{E}_w) + \epsilon)$, where $\mathbf{E}_w$ are all embeddings of word $w$.

\textbf{4. Statistical tests:} We compute Spearman correlation between: Frequency and polysemy (Martin's Law) and frequency and specificity (frequency-specificity tradeoff). We focus on the top 500 most frequent words to ensure reliable clustering and reduce computational cost.

\subsection{Evaluation Metrics}

\begin{itemize}
    \item \textbf{Spearman $\rho$}: Rank correlation between frequency and polysemy/specificity
    \item \textbf{Polysemous word count}: Number of words with $>1$ cluster
    \item \textbf{Mean polysemy}: Average number of senses per word
    \item \textbf{Semantic differentiation}: Mean polysemy as a proxy for semantic structure richness
\end{itemize}

\begin{figure}[t]
    \centering
    \includegraphics[width=\linewidth]{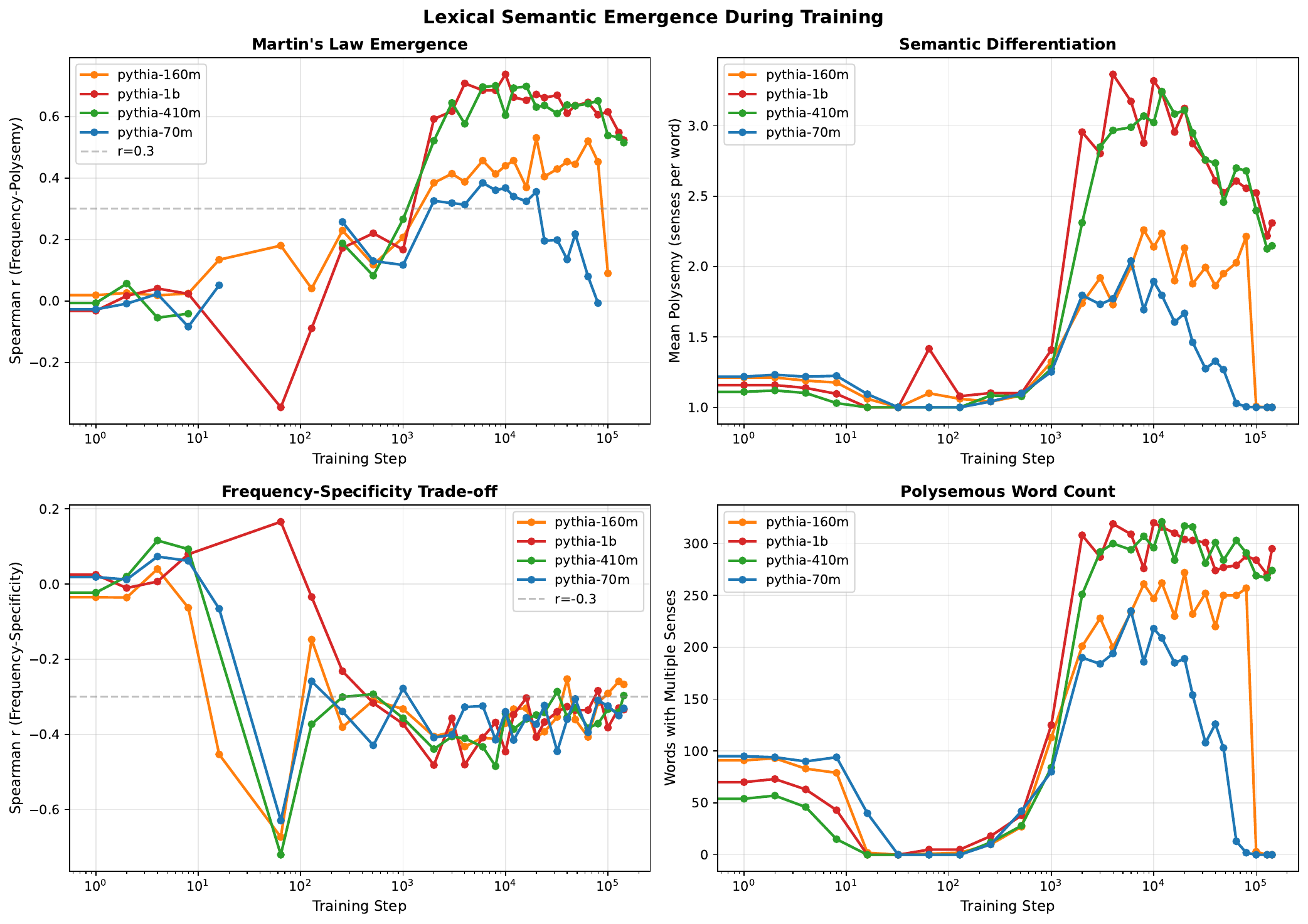}
    \caption{\textbf{Semantic emergence across training.} \textit{Top-left:} Martin's Law (frequency-polysemy correlation) shows non-monotonic trajectory with peak at $\sim10^4$ steps. \textit{Top-right:} Mean polysemy (semantic differentiation) collapses in small models at late checkpoints. \textit{Bottom-left:} Frequency-specificity tradeoff remains stable across training. \textit{Bottom-right:} Polysemous word count diverges by model scale, with catastrophic collapse in small models.}
    \label{fig:emergence}
\end{figure}

\section{Results}

\subsection{Developmental Trajectory of Martin's Law}

Figure \ref{fig:emergence} shows the evolution of Martin's Law across training. All models exhibit a consistent three-phase pattern:

\textbf{Phase 1: Emergence (cp 0--100):} Initially, the frequency-polysemy correlation is near zero. Around checkpoint 100, all models show rapid increase, indicating the emergence of differentiated semantic structure.

\textbf{Phase 2: Peak (cp $10^3$--$10^4$):} Martin's Law reaches maximum strength at checkpoint $\sim10^4$, with Spearman $\rho > 0.6$ for all models. This represents a balanced zone between the extremes where semantic organization most closely mirrors natural language patterns.

\textbf{Phase 3: Degradation (cp $10^4$--$10^5$):} After the peak, the correlation strength declines. The larger models (1B, 410M) gracefully degrade to $\rho \approx 0.5$, while the smaller models (160M, 70M) collapse toward zero.

\subsection{Catastrophic Semantic Collapse in Small Models}

Small models exhibit a striking failure mode at late checkpoints. For models 70M and 160M, polysemous word count drops to zero around checkpoint $10^5$ (Figure \ref{fig:emergence}, bottom-right panel). This is accompanied by semantic differentiation breakdown (Figure \ref{fig:emergence}, top-right panel), where mean polysemy collapses.

Critically, this is not merely cluster merging: the models are generating semantically impoverished text where words lack contextual variation. In contrast, the 1B model maintains $\sim275$ polysemous words even at late checkpoints, though with weakened Martin's Law correlation.

\subsection{Stable Frequency-Specificity Tradeoff}

Unlike Martin's Law, the frequency-specificity tradeoff remains remarkably stable: Spearman $\rho \approx -0.4$ across all models and checkpoints later than $ \sim 10^3$ (Figure \ref{fig:emergence}, bottom-left panel). This is weaker than typical natural language values ($\rho \approx -0.5$ to $-0.7$), suggesting LLM-generated text exhibits a "flatter" semantic space where frequency less strongly predicts generality.

\subsection{Model Scale and Semantic Capacity}

Larger models consistently exhibit higher polysemous word counts throughout training (Figure \ref{fig:emergence}, bottom-right panel). The 1B and 410M models sustains $\sim275$ to $300$ polysemous words at late checkpoints, the smaller models between $\sim200$ and $250$. This suggests a capacity threshold: models below $\sim200$M parameters cannot maintain diverse semantic representations under continued training.

\section{Discussion}

\subsection{Non-Monotonic Semantic Development}

Our central finding is that Martin's Law compliance in LLM-generated text is \textit{not monotonically increasing with training}. Instead, there exists an optimal intermediate checkpoint ($\sim10^4$ steps) where semantic organization best reflects natural language structure. This challenges the implicit assumption that longer training produces more human-like linguistic properties.

The degradation after peak suggests competing pressures in late training: memorization may override the distributional semantic structure that gives rise to Martin's Law. Alternatively, models may be collapsing into degenerate solutions that satisfy training objectives while sacrificing semantic richness.

\subsection{Capacity Thresholds for Semantic Maintenance}

The catastrophic collapse in small models reveals a critical capacity threshold. Below $\sim200$M parameters, models cannot sustain polysemous representations under continued training. This may reflect fundamental limits on how much semantic structure can be compressed into limited parameter space when simultaneously optimizing for next-token prediction.

Intriguingly, the 1B model's polysemous word count remains stable relatively even as Martin's Law weakens ($\rho: 0.6 \to 0.5$). This suggests semantic \textit{reorganization} rather than collapse: polysemy persists but becomes less frequency-governed.

\subsection{Weaker Frequency-Specificity Tradeoff}

The stable but weak frequency-specificity correlation ($\rho \approx -0.3$) suggests LLM-generated text occupies a semantically "flatter" space than natural language. High-frequency words may not be as consistently general, or low-frequency words not as consistently specific, as in human-produced text. This could indicate fundamental differences in how LLMs allocate semantic content across the frequency spectrum.

\subsection{Implications for LLM Evaluation}

Our findings suggest a novel evaluation paradigm: testing whether model-generated text respects established linguistic laws. This complements existing approaches (perplexity, downstream tasks) by directly probing emergent linguistic structure. The non-monotonic Martin's Law trajectory suggests that checkpoint selection matters---models at intermediate training may produce more linguistically natural text than fully trained models.

\section{Limitations and Future Work}

\subsection{Methodological Refinements}

\textbf{Clustering methods and parameters:} Our use of DBSCAN with fixed $\epsilon=0.3$ represents one point in the parameter space. Systematic sensitivity analysis across $\epsilon$ values (e.g., 0.2--0.5) would establish robustness. Additionally, comparing DBSCAN results with alternative clustering methods (agglomerative clustering with various linkage criteria, Gaussian Mixture Models with automatic component selection, or affinity propagation) would validate that our findings are not artifacts of a particular clustering algorithm. Hierarchical clustering approaches could also reveal whether sense granularity changes across training checkpoints.

\textbf{Sample size and statistical power:} We analyze 100 samples (51,200 tokens) per checkpoint. While sufficient for detecting large effects, this limits our ability to reliably estimate polysemy for mid-frequency words. Future work should generate 500--1,000 samples per checkpoint to improve frequency estimates and reduce sampling noise, particularly for the critical $10^3$--$10^5$ checkpoint range where semantic reorganization occurs.

\subsection{Comparison with Human-Written Text}

A critical gap in our current work is the absence of direct comparison with human-written text. We observe Martin's Law emerging and degrading in LLM-generated text, but without a human baseline from the same domain, we cannot determine whether the peak correlation ($\rho \approx 0.6$) represents successful learning of human-like semantic structure or merely a coincidental pattern.

\textbf{Training corpus baseline:} Future work should analyze text from the Pile (or successor corpora like RedPajama or Dolma) using the same embedding extraction and clustering pipeline. This would establish: (1) what Martin's Law correlation exists in human text when measured via our methodology; (2) whether models at checkpoint $10^4$ genuinely replicate human semantic structure or exhibit qualitatively different patterns; (3) which specific words are polysemous in human vs. LLM text, revealing whether models develop appropriate vs. spurious polysemy.

\textbf{Prompted generation:} Our current approach uses unconditional or minimally prompted generation, which may artificially reduce semantic richness compared to human text produced in natural discourse contexts. A three-way comparison design would be revealing:
\begin{enumerate}
    \item \textit{Human-written text:} Original documents from training corpora
    \item \textit{Prompted generation:} LLM continuations from real document headlines/openings
    \item \textit{Unconditional generation:} Our current approach
\end{enumerate}
This design would isolate whether differences in Martin's Law arise from generation vs. human authorship, or from lack of realistic discourse framing. We hypothesize that prompted generation would show stronger Martin's Law (closer to human text) because realistic context constrains word usage toward appropriate polysemy patterns.

\subsection{Cross-Model and Cross-Linguistic Extensions}

\textbf{Other model families:} We study Pythia (GPT-NeoX architecture). Extending to other models with public checkpoints like OLMo \cite{groeneveld2024olmo}, Amber, or Cerebras-GPT would test whether our findings reflect general LLM training dynamics or architecture-specific phenomena. Comparing models trained on different corpora (Pile vs. RedPajama vs. Dolma) would reveal whether Martin's Law trajectories depend on training data properties.

\textbf{Multilingual analysis:} Testing Martin's Law in LLM-generated text across languages would provide strong evidence for universal vs. language-specific patterns. Corpora like ROOTS (used for BLOOM) offer multilingual training data, enabling comparison of semantic emergence across typologically diverse languages.

\subsection{Other Linguistic Regularities}

This work focuses on Martin's Law. A comprehensive evaluation of linguistic law compliance should include:

\begin{itemize}
    \item \textbf{Zipf's Law:} Word frequency distributions in generated vs. human text
    \item \textbf{Heaps' Law:} Vocabulary growth rates as a function of text length
    \item \textbf{Menzerath-Altmann Law:} Relationship between construct size and constituent size
    \item \textbf{Brevity law:} Frequency-length correlations
    \item \textbf{Syntactic complexity:} Dependency length, phrase structure depth
\end{itemize}

Understanding which laws emerge, when, and which fail would provide a comprehensive picture of linguistic naturalism in LLM-generated text.

\subsection{Mechanistic Understanding}

Our results are descriptive. Controlled interventions focused on causal mechanisms could reveal \textit{why} Martin's Law emerges and degrades: Does the learning rate scheduling affect the semantic peak location? Do models trained with different objectives (masked LM vs. causal LM) show different trajectories? Can we identify which layers/attention heads are responsible for polysemous representations and does the degradation phase reflect overfitting, mode collapse, or memorization?

Artificially enhancing or suppressing polysemy during training (e.g., via contrastive objectives or sense-aware losses) could test whether Martin's Law compliance improves downstream task performance or linguistic naturalness.

\section{Conclusion}

We present the first investigation of Martin's Law in LLM-generated text, revealing that compliance with this fundamental linguistic regularity follows a non-monotonic trajectory during training. Our findings establish that: (1) semantic structure in LLM-generated text peaks at intermediate checkpoints rather than improving monotonically; (2) model capacity determines whether semantic richness can be maintained under continued training; (3) testing linguistic laws provides a powerful lens for understanding emergent properties of neural language models.

This work opens a new research direction: systematic evaluation of whether LLM-generated text respects the statistical regularities that characterize human language. As LLMs become increasingly central to language technology, understanding the linguistic structure of their outputs, not just their task performance, becomes critical.

\bibliographystyle{plain}

\end{document}